\documentclass[10pt,twocolumn,letterpaper]{article}

\usepackage[pagenumbers]{cvpr} 
\usepackage{graphicx}
\usepackage{xcolor}
\usepackage{multirow}
\usepackage{comment}

\usepackage{booktabs}
\usepackage{xcolor}

\newcommand{\rev}[1]{{\color{blue}#1}}

\definecolor{cvprblue}{rgb}{0.21,0.49,0.74}
\usepackage[pagebackref,breaklinks,colorlinks,allcolors=cvprblue]{hyperref}

\usepackage[margin=1in]{geometry}
\usepackage{tikz}
\usepackage{subcaption}
\usetikzlibrary{shapes.geometric, positioning, matrix}
\usepackage{amsmath}
\usepackage{pgfplots}

\def\confName{CVPR}
\def\confYear{2027}

\title{ReACE: Revisiting Local Descriptors for Accelerated Coordinate Encoding
}

\author{Yahav Vinokur \qquad Yoli Shavit\\
Faculty of Engineering, Bar-Ilan University, Ramat Gan, Israel\\
}

\begin{document}
\maketitle
\begin{abstract}
Scene coordinate regression offers an efficient alternative to hierarchical feature-matching pipelines for visual relocalization by predicting 3D scene coordinates directly from image pixels. Accelerated Coordinate Encoding (ACE) made this practical for on-site mapping by training a compact per-scene regressor over shuffled patch features extracted from a pretrained vision encoder. In this work, we present ReACE, which revisits the use of local descriptors within the ACE paradigm. We investigate how local descriptors can help address the spatial coarseness of foundation backbones by providing both a salient keypoint sampling guide and an informative point-level representation. ReACE samples DINOv2 features at detected keypoints and concatenates them with local descriptors to train a compact regression head. This strategy preserves the rapid mapping and minimal memory footprint of the ACE framework while improving localization accuracy. Across standard indoor and outdoor benchmarks, ReACE outperforms existing scene coordinate regressors with a comparable mapping budget and achieves accuracy competitive with heavy matching pipelines. Through extensive ablations, we analyze the individual contributions of keypoint-guided sampling and descriptor fusion to these accuracy gains. Code is available at: \url{https://github.com/yahav6893/ReACE}.
\end{abstract}
\section{Introduction}
\label{sec:intro}
Estimating the camera pose of a query image within a mapped environment is central to robotics and autonomous navigation. Hierarchical localization pipelines solve this using Perspective-n-Point (PnP) inside a RANSAC loop~\cite{fischler1981random} by building 2D-to-3D correspondences across the scene~\cite{sarlin2019coarse,sarlin2020superglue,taira2018inloc}. This is typically achieved in two steps, first retrieving co-visible reference images and then matching local query features against reference keypoints anchored to a prebuilt 3D point cloud. Although these pipelines deliver state-of-the-art precision, constructing a 3D map requires hours of compute and storing dense descriptor sets incurs gigabyte-scale memory footprints. Direct alternatives like absolute pose regression bypass explicit 3D models entirely by regressing camera pose directly from the query image~\cite{kendall2015posenet,shavit2021multiscene}, yet they lack geometric constraints and remain far less accurate.

Scene coordinate regression resolves this trade-off by predicting 3D scene coordinates directly from 2D pixels, establishing correspondences for PnP-RANSAC to achieve competitive pose accuracy without requiring explicit 3D models or feature storage~\cite{brachmann2017dsac,brachmann2021dsacstar}. Accelerated Coordinate Encoding (ACE)~\cite{brachmann2023accelerated} made mapping practical in minutes by training a compact per-scene Multi-Layer Perceptron (MLP) on large, decorrelated buffers of patch features extracted from a frozen image encoder. Subsequent efforts to improve accuracy introduced extensive machinery around this core design, including global scene embeddings~\cite{wang2024glace} and multi-scene pretraining regimes~\cite{bruns2025ace}.

\begin{figure*}[t]
  \centering
  \includegraphics[scale=0.84]{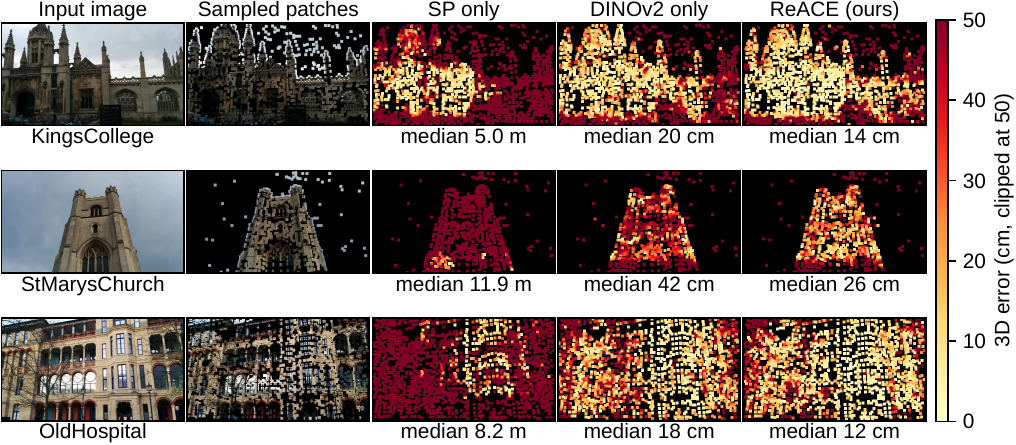}
  \caption{
  {\textbf{Per-keypoint 3D error when using different inputs to the SCR head.} Each patch is colored by the distance between its sampled keypoint and the ground-truth viewing ray (unsampled patches appear in black). Using the local descriptor (SuperPoint; SP) alone achieves accuracy only on locally distinctive structures, whereas backbone features (DINOv2) alone remain spatially coarse and exhibit localized failure regions. ReACE fuses both representations, consistently achieving superior accuracy across the entire scene. Measurement details are provided in the supplementary material {(Sec.~S4)}.}
  }
  \label{fig:teaser}
\end{figure*}

Recently, experimental results reported by Bruns~\etal~\cite{bruns2025ace} showed that simply replacing ACE's convolutional backbone with DINOv2 yields substantial accuracy gains. Yet while DINOv2 provides robust semantic representations that mitigate perceptual aliasing, its tokens are extracted at coarse patch granularity, potentially limiting fine spatial precision. We revisit this representation design from the perspective of hierarchical localization pipelines, which rely on local features sampled and optimized specifically for spatial repeatability and fine geometric discriminativeness. We hypothesize that these local descriptors can resolve the spatial coarseness of foundation backbones by serving simultaneously as a salient sampling guide and as a complementary point-level representation. Building on this hypothesis, we propose ReACE, which \textit{revisits} local descriptors within the ACE framework by sampling DINOv2 at keypoint locations and concatenating patch features with point-level descriptors, improving localization accuracy while preserving ACE's short mapping time and megabyte-scale footprint (\cref{fig:teaser}).

Across standard outdoor and indoor localization benchmarks, ReACE outperforms state-of-the-art scene coordinate regressors with a comparable mapping budget despite its simpler design, while closing the gap to heavy feature-matching pipelines and in several cases exceeding them. Through controlled ablations, we isolate the origin of these gains. We show that performance improvements stem from the point-level geometric information injected by local descriptors rather than increased capacity alone, with consistent gains across distinct descriptor families. In addition, our proposed direct concatenation is competitive with heavier residual or attention-based fusion variants. Multi-seed controls further confirm that the margins over DINO-ACE exceed run-to-run variance by more than an order of magnitude. Finally, incorporating multiple complementary descriptors offers additional performance gains on some benchmarks at a modest increase in model footprint.

In summary, our contributions are as follows.
\begin{itemize}
\item We revisit scene coordinate regression from the perspective of hierarchical matching pipelines and introduce ReACE, which leverages local keypoint descriptors to provide both a salient sampling guide and point-level geometric specificity that complement coarse semantic patch representations from vision foundation backbones.
\item ReACE outperforms state-of-the-art scene coordinate regressors with a comparable mapping budget across standard indoor and outdoor benchmarks and narrows the accuracy gap to heavy feature-matching pipelines while requiring orders of magnitude less storage and mapping time.
\item Through extensive controlled ablations and multi-seed analysis, we demonstrate that gains stem from point-level geometric information rather than added capacity, with consistent improvements across diverse descriptor families, superior efficiency for direct concatenation over complex fusion variants, and stable performance well exceeding run-to-run variance.
\end{itemize}

\section{Related Work}
\label{sec:related_work}
Structure-based visual relocalization methods~\cite{sattler2016efficient} estimate camera pose from 2D-to-3D correspondences using PnP inside a RANSAC loop. Unlike absolute or relative pose regression~\cite{dong2025reloc3r,shavit2021multiscene,kendall2015posenet}, which predict camera coordinates directly from images in a forward pass, structure-based methods enforce geometric consistency and yield higher accuracy~\cite{sattler2019understanding}. This family primarily includes hierarchical feature matching pipelines and scene coordinate regression.

\subsection{Hierarchical Feature Matching Pipelines}
\label{sec:rw_matching}
Hierarchical localization systems~\cite{taira2018inloc,sarlin2019coarse} follow a coarse-to-fine pipeline. A query image is first encoded into a global descriptor for visual place recognition, retrieving a candidate subset of co-visible reference views from a database. Next, matching local features between the query and retrieved reference images yields 2D-to-2D correspondences that are paired with 3D points from a prebuilt structure-from-motion reconstruction~\cite{schoenberger2016colmap}. While hierarchical pipelines achieve state-of-the-art accuracy across challenging conditions, constructing the 3D map requires hours of computation, and storing reference keypoints alongside dense descriptor pools incurs gigabyte-scale memory footprints.

\subsection{Scene Coordinate Regression}
\label{sec:rw_scr}
Scene coordinate regression (SCR) eliminates the need to store local descriptors and an explicit 3D model by training a neural network to regress 3D scene coordinates directly from 2D pixel observations, retaining pose estimation through a robust PnP solver. Differentiable RANSAC formulations introduced by DSAC~\cite{brachmann2017dsac,brachmann2021dsacstar} enabled end-to-end learning under a task loss that minimizes camera reprojection error. Accelerated Coordinate Encoding (ACE)~\cite{brachmann2023accelerated} substantially improved training efficiency, reducing mapping times to minutes and storage to several megabytes by optimizing a compact scene-specific MLP with a reprojection loss over decorrelated patch features extracted from a frozen convolutional backbone pretrained with direct depth supervision.

Subsequent work sought to mitigate perceptual aliasing and improve accuracy by introducing heavier architectural components around this core setup. GLACE~\cite{wang2024glace} conditions the coordinate head on a global image vector, requiring an auxiliary encoder and a longer mapping budget. ACE-G~\cite{bruns2025ace} pretrains an expressive transformer head across thousands of scenes via an extensive multi-stage curriculum. Rather than modifying the head, Bruns et al.~\cite{bruns2025ace} showed that simply replacing ACE's convolutional backbone with frozen DINOv2 patch features yields substantial accuracy gains while maintaining simple per-scene training.

\subsection{Local Descriptors in Visual Localization}
\label{sec:rw_features}
Local descriptors characterize the visual region around a keypoint, designed to remain matchable under viewpoint and illumination changes. Modern learned features~\cite{detone2018superpoint,revaud2019r2d2,tyszkiewicz2020disk,zhao2023aliked,edstedt2024dedode,potje2024xfeat} form the foundation of hierarchical pipelines, where descriptor choice governs map storage, matching speed, and localization accuracy.

Sparse keypoint mechanisms have also emerged within coordinate regression. Inspired by FocusTune~\cite{nguyen2024focustune}, SeqACE~\cite{xu2025seqace} distills a SuperPoint~\cite{detone2018superpoint} keypoint detector into a head on the ACE backbone to restrict regression sampling to salient regions, yet it discards the associated local descriptors. D2S~\cite{bui2024d2s} regresses scene coordinates directly from local descriptors, bypassing dense features from vision foundation backbones entirely. Each descriptor attends to all others in the image through graph-attention layers and is supervised by a pre-built SfM model of the scene, requiring roughly nine hours of training per scene. In contrast, ReACE fuses local descriptors with backbone features and learns from posed images alone within minutes. 

To overcome perceptual aliasing across multi-room or city-scale scenes, recent extensions pair sparse descriptors with global scene contexts. R-SCoRe~\cite{jiang2025rscore} conditions descriptor-based regression on learned global embeddings, dynamically scaling the MLP capacity with the image set size. Most closely related to our work, Nguyen et al.~\cite{nguyen2026robust} fuse frozen DeDoDe descriptors with DINOv2 features. However, their architecture addresses large-scale place disambiguation by training a NetVLAD-inspired aggregator over an image covisibility graph, compressing DINOv2 features into a single 256-dimensional global image fingerprint that is broadcast across all keypoints. Furthermore, their formulation scales the MLP width with scene size to dozens of MB and relies on multi-hypothesis retrieval during inference under severe aliasing. R-SCoRe and~\cite{nguyen2026robust} evaluate only on large-scale scenes (Aachen, Hyundai Department Store) and train for hours per scene, so neither reports on our evaluated benchmarks.
In contrast, ReACE does not require covisibility graph, aggregator pretraining, or multi-hypothesis inference, and maps a scene from posed images alone in roughly five minutes and a few megabytes. In addition, while coarse-to-fine feature pairing has proven effective for dense correspondence estimation in RoMa~\cite{edstedt2024roma}, combining point-wise foundation features with local descriptors was not yet efficiently demonstrated for scene coordinate regression.
\section{Method}
\label{sec_method}
\begin{figure*}[t]
  \centering
  \begin{subfigure}{\textwidth}
    \centering
    \includegraphics[scale=0.52]{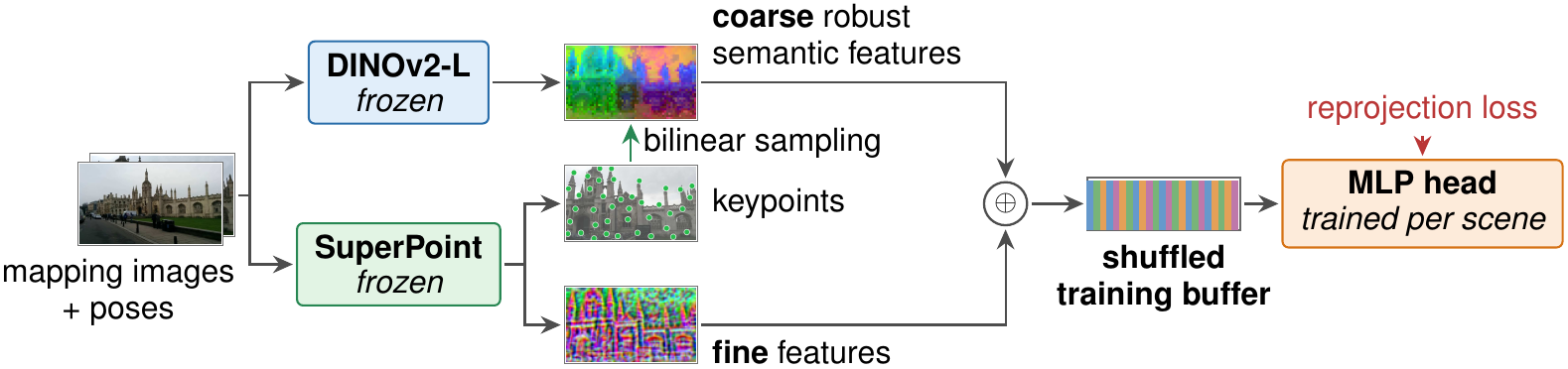}
    \caption{Mapping: once per scene.}
    \label{fig:method_mapping}
  \end{subfigure}\\[1.5mm]
  \begin{subfigure}{\textwidth}
    \centering
    \includegraphics[scale=0.52]{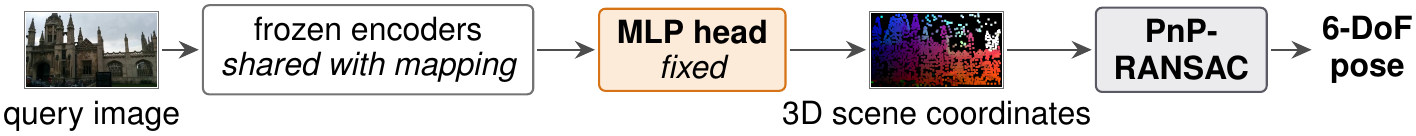}
    \caption{Relocalization: per query.}
    \label{fig:method_reloc}
  \end{subfigure}
  \caption{\textbf{Overview of ReACE.} Backbone features are sampled at detected keypoint locations, concatenated with local descriptors, and fed into a lightweight per-scene MLP. (a) During mapping, fused per-keypoint representations from all training images populate a shuffled buffer used to optimize the scene head under ACE's reprojection loss. (b) During localization, the same frozen feature extractor feeds the trained head to predict 3D scene coordinates for robust PnP pose estimation.}
  \label{fig:method}
\end{figure*}

We begin with the technical formulation of Accelerated Coordinate Encoding (ACE)~\rev{\cite{brachmann2023accelerated}} to establish our baseline pipeline. Building on this foundation, we then introduce ReACE, detailing our keypoint-guided sampling, descriptor fusion, and {pose refinement}. \Cref{fig:method} illustrates the ReACE architecture alongside its training and inference pipelines.

\subsection{Accelerated Coordinate Encoding (ACE)}
\label{sec:method_ace}

Given mapping images $\{I_j\}$ with known camera poses $\{\mathbf{T}_j\}$ and intrinsics $\{\mathbf{K}_j\}$, ACE~\cite{brachmann2023accelerated} performs scene coordinate regression using a frozen, scene-agnostic encoder for dense feature extraction and a compact scene-specific MLP head $h$ that maps each local feature $\mathbf{f}_i$ at pixel location $\mathbf{x}_i$ to its 3D scene coordinate $\mathbf{y}_i = h(\mathbf{f}_i)$. During training, the head $h$ is optimized to minimize the robust reprojection error $\sum_i \rho(\pi_j(\mathbf{y}_i) - \mathbf{x}_i)$, where $\pi_j$ denotes the camera projection using $(\mathbf{T}_j, \mathbf{K}_j)$ and $\rho$ is a dynamically clamped loss function.

Training batches are drawn uniformly at random from a pre-extracted buffer of patch features pooled across all mapping images. This cross-image shuffling decorrelates intra-batch gradients and enables full convergence within minutes. At inference time, 3D coordinates are predicted across the query image $I_q$, and the 6-Degrees-of-Freedom (6-DoF) camera pose is recovered from the resulting 2D-to-3D correspondences using PnP inside a RANSAC loop. Because $h$ evaluates each point independently, any modification confined to the per-point feature vector seamlessly preserves this fast buffer-training pipeline. We build on the DINO-ACE formulation~\cite{bruns2025ace}, which replaces the original convolutional backbone with a frozen DINOv2 vision foundation model~\cite{oquab2023dinov2}.

\subsection{{ReACE: Local Descriptors in ACE}}
\label{sec_method_reace}

ReACE extends DINO-ACE by altering where features are sampled, augmenting the per-point input with local descriptors, and {tightening the pose refinement threshold} (\cref{fig:method}). All remaining architecture, training, and inference mechanics are inherited from ACE and DINO-ACE.

\paragraph{Keypoint-guided Sampling.}
Instead of sampling DINOv2 patch features across a dense grid, ReACE samples features only at keypoint locations. An off-the-shelf detector such as SuperPoint~\cite{detone2018superpoint} extracts a sparse set of 2D keypoints per image, and the backbone representation at each location is bilinearly interpolated from the four neighboring patch features.

\paragraph{Descriptor Fusion.}
For each keypoint $\mathbf{x}_k$ with local descriptor $\mathbf{d}_k$, the input to the regression head is formed by concatenating the interpolated backbone feature with the scaled descriptor,
\begin{equation}
  \mathbf{y}_k = h\left(\left[\,\mathbf{f}(\mathbf{x}_k),\, s\,\mathbf{d}_k\,\right]\right),
  \label{eq_fusion}
\end{equation}
where $\mathbf{f}(\mathbf{x}_k)$ denotes the bilinearly interpolated backbone feature and $s$ is a fixed scaling factor. Features and descriptors are extracted once during buffer initialization, so their runtime cost is fully reflected in our reported mapping budgets. Because the local descriptor is $\ell_2$-normalized across its $D$ dimensions, its per-element magnitude is $1/\sqrt{D}$, which falls well below that of the backbone features. Setting $s = \sqrt{D}$ brings the two streams to comparable per-element scales, effectively fixing the total norm of the descriptor stream to $s$. The coordinate head retains ACE's MLP architecture, modified only by widening the input layer to receive the concatenated dimension. Since the fused representation remains strictly pointwise, ACE's shuffled buffer optimization transfers to ReACE. While the default ReACE formulation uses a single feature, we also consider a variant of this formulation, termed ReACE-MF (multi-feature). ReACE-MF  concatenates additional per-point features into the same input vector, where all descriptors are jointly $\ell_2$-normalized before applying $s$. Concrete model variants and hyperparameter choices are summarized in \cref{sec:exp_setup} and Sec.~S4.

\paragraph{Camera Pose Estimation.}
At inference time, ReACE extracts and fuses features at detected keypoints to predict sparse 2D-to-3D correspondences. The camera pose is estimated with PnP inside a RANSAC loop, followed by iterative refinement of the pose on its inliers~\cite{brachmann2021dsacstar}. DSAC* and ACE use the same 10\,px inlier threshold for both steps. We keep it for the search but refine with a tighter threshold, which improves precision (\cref{sec:ablations}).

The individual impact of keypoint-guided sampling, descriptor fusion , and the tight refinement threshold is analyzed in our ablation studies (\Cref{tab:abl_components}). 

\begin{table*}[t]
\centering
\scriptsize
\setlength{\tabcolsep}{3pt}
\caption{{\textbf{Quantitative results on Cambridge Landmarks.}} We report the median translation (cm) and rotation ($^\circ$) errors, highlighting the best SCR entry per budget group in bold. {Baseline performance is quoted from the original publications (HLoc as reported in~\cite{brachmann2023accelerated}); methods are grouped by approximate mapping budget ({Table.~S10}).}}
\label{tab:main_cambridge}
\begin{tabular}{llcccccc}
\toprule
Method & Size & GreatCourt & KingsCollege & OldHospital & ShopFacade & StMarysChurch & Avg. \\
\midrule
\multicolumn{8}{l}{{\textit{Non-SCR}}} \\
HLoc (SP+SG)~\cite{sarlin2019coarse,sarlin2020superglue} & $\sim$800\,MB
  & 16 / 0.1 & 12 / 0.2 & 15 / 0.3 & 4 / 0.2 & 7 / 0.2 & 10.8 / 0.20 \\
\midrule
\multicolumn{8}{l}{{\textit{SCR, $>$5\,h budget}}} \\
{D2S~\cite{bui2024d2s}} & {22\,MB}
  & {23 / 0.11} & {7 / 0.12} & {15 / 0.29} & {3 / 0.17} & {8 / 0.25} & {11.2 / 0.19} \\
\midrule
\multicolumn{8}{l}{\textit{SCR, {$\sim$25-minute} budget}} \\
ACE-Poker ($\times$4)~\cite{brachmann2023accelerated} & 16\,MB
  & 28 / 0.1 & 18 / 0.3 & 25 / 0.5 & 5 / 0.3 & \textbf{9} / \textbf{0.3} & 17.0 / 0.30 \\
HSeq-Ensemble ($\times$4)~\cite{xu2025seqace} & 16.4\,MB
  & 27 / 0.1 & 17 / 0.3 & 24 / 0.5 & 5 / 0.3 & \textbf{9} / \textbf{0.3} & 16.4 / 0.30 \\
GLACE~\cite{wang2024glace} & 13\,MB
  & \textbf{19} / 0.1 & 19 / 0.3 & \textbf{17} / 0.4 & \textbf{4} / \textbf{0.2} & \textbf{9} / \textbf{0.3} & 13.6 / 0.26 \\
ACE-G {(25-min)}~\cite{bruns2025ace} & 12\,MB
  & 50.6 / 0.3 & 21.1 / 0.3 & 22.5 / 0.5 & 7.4 / 0.3 & 24.9 / 0.8 & 25.3 / 0.44 \\
ReACE-Poker (ours; $\times$4) & 25.1\,MB
  & 21.6 / \textbf{0.09} & \textbf{10.7} / \textbf{0.17} & 17.2 / \textbf{0.35}
  & 5.7 / 0.25 & 10.8 / 0.34 & \textbf{13.2} / \textbf{0.24} \\
\midrule
\multicolumn{8}{l}{\textit{SCR, {$\sim$5-minute} budget}} \\
ACE~\cite{brachmann2023accelerated} & 4\,MB
  & 43 / 0.2 & 28 / 0.4 & 31 / 0.6 & 5 / 0.3 & 18 / \textbf{0.6} & 25.0 / 0.42 \\
DINO-ACE~\cite{bruns2025ace} & 4\,MB
  & 61.4 / 0.3 & 38.9 / 0.4 & 39.8 / 0.6 & 5.3 / 0.3 & 74 / 1.8 & 43.9 / 0.68 \\
HSeq-Single~\cite{xu2025seqace} & 4.1\,MB
  & 38 / 0.2 & 25 / 0.4 & 27 / 0.6 & 5 / 0.3 & \textbf{17} / \textbf{0.6} & 22.4 / 0.42 \\
ACE-G {(5-min)}~\cite{bruns2025ace} & 12\,MB
  & 68.2 / 0.4 & 26.3 / 0.4 & 28.7 / 0.5 & 7.8 / 0.3 & 44.2 / 1.4 & 35.0 / 0.60 \\
ReACE (ours) & 6.3\,MB
  & \textbf{24.4} / \textbf{0.11} & \textbf{14.9} / \textbf{0.20} & \textbf{19.6} / \textbf{0.38}
  & \textbf{4.1} / \textbf{0.18} & 34.0 / 1.02 & \textbf{19.4} / \textbf{0.38} \\
\bottomrule
\end{tabular}

\end{table*}

\begin{table*}[t]
\centering
\scriptsize
\setlength{\tabcolsep}{2pt}
\caption{\textbf{Quantitative results on Indoor-6.} We report recall (\%) under (25\,cm, 5$^\circ$), the median translation (cm) and rotation ($^\circ$) errors, highlighting the best SCR entry per budget group in bold. {Baseline performance is quoted from~\cite{bruns2025ace}, and D2S from~\cite{bui2024d2s}, which reports recall only at (5\,cm, 5$^\circ$) and whose average we recompute. Methods are grouped by approximate mapping budget ({Table~S10}).}}
\label{tab:main_indoor6}
\begin{tabular}{llccccccc}
\toprule
Method & Size & Scene 1 & Scene 2a & Scene 3 & Scene 4a & Scene 5 & Scene 6 & Avg. \\
\midrule
\multicolumn{9}{l}{\textit{Non-SCR}} \\
Reloc3r~\cite{dong2025reloc3r} & $\sim$500\,MB
  & 76 / 8 / 0.6 & 76 / 10 / 0.9 & 87 / 4 / 0.5 & 82 / 9 / 0.8 & 50 / 22 / 1.4 & 86 / 3 / 0.6 & 76 / 9 / 0.8 \\
MASt3R~\cite{leroy2024mast3r}+Kapture & $\sim$5\,GB
  & 92 / 2.4 / 0.4 & 97 / 2.6 / 0.3 & 96 / 1.8 / 0.4 & 96 / 2.6 / 0.5 & 85 / 4.3 / 0.7 & 95 / 1.3 / 0.3 & 94 / 2.5 / 0.5 \\
\midrule
\multicolumn{9}{l}{{\textit{SCR, $>$5\,h budget}}} \\
{D2S~\cite{bui2024d2s}} & {22\,MB}
  & {-- / 4.8 / 0.81} & {-- / 4.0 / 0.41} & {-- / 3.6 / 0.69} & {-- / 2.1 / 0.48} & {-- / 5.8 / 0.90} & {-- / 2.4 / 0.48} & {-- / 3.78 / 0.63} \\
\midrule
\multicolumn{9}{l}{\textit{SCR, {$\sim$25-minute} budget}} \\
GLACE~\cite{wang2024glace} & 6\,MB
  & 90 / 3.8 / 0.7 & \textbf{100} / 3.9 / 0.4 & 93 / 2.9 / 0.6 & 99 / 2.3 / 0.5 & 79 / 6.1 / 0.9 & \textbf{97} / 2.1 / 0.4 & 93 / 3.5 / 0.6 \\
ACE-G {(25-min)}~\cite{bruns2025ace} & 12\,MB
  & 96 / 3.5 / 0.6 & 98 / 3.9 / 0.4 & 98 / 3.8 / 0.8 & 99 / 3.2 / 0.8 & 97 / 5.1 / 0.8 & 91 / 2.7 / 0.6 & 96 / 3.7 / 0.7 \\
ReACE-Poker (ours; $\times$4) & 25.1\,MB
  & \textbf{{97.8}} / \textbf{{1.99}} / \textbf{{0.34}} & {99.4} / \textbf{{2.66}} / \textbf{{0.26}} & \textbf{{98.5}} / \textbf{{1.68}} / \textbf{{0.32}}
  & \textbf{100.0} / \textbf{{1.75}} / \textbf{{0.41}} & \textbf{{98.1}} / \textbf{{2.55}} / \textbf{0.42} & {96.4} / \textbf{{1.13}} / \textbf{0.24}
  & \textbf{{98.4}} / \textbf{{1.96}} / \textbf{{0.33}} \\ 
\midrule
\multicolumn{9}{l}{\textit{SCR, {$\sim$5-minute} budget}} \\
ACE~\cite{brachmann2023accelerated} & 4\,MB
  & 52 / 17.7 / 2.4 & 86 / 8.0 / 0.8 & 66 / 9.0 / 1.6 & 94 / 4.6 / 0.9 & 51 / 21.6 / 4.0 & 70 / 5.2 / 0.9 & 70 / 11.0 / 1.8 \\
DINO-ACE~\cite{bruns2025ace} & 4\,MB
  & 90 / 5.4 / 0.9 & 97 / 5.2 / 0.6 & 91 / 6.0 / 1.2 & 96 / 4.7 / 1.2 & 89 / 7.5 / 1.2 & 85 / 4.8 / 1.0 & 91 / 5.6 / 1.0 \\
ACE-G {(5-min)}~\cite{bruns2025ace} & 12\,MB
  & 94 / 4.3 / 0.7 & 98 / 4.8 / \textbf{0.4} & 96 / 4.7 / 0.9 & 98 / 3.7 / 0.9 & 93 / 5.9 / 1.0 & 92 / 3.9 / 0.8 & 95 / 4.5 / 0.8 \\
ReACE (ours) & 6.3\,MB
  & \textbf{95.9} / \textbf{2.91} / \textbf{0.50} &  \textbf{98.2} / \textbf{3.82} / \textbf{0.4}
  & \textbf{96.9} / \textbf{2.97} / \textbf{0.57} & \textbf{99.8} / \textbf{2.24} / \textbf{0.51}
  & \textbf{96.3} / \textbf{3.99} / \textbf{0.64} & \textbf{95.1} / \textbf{1.78} / \textbf{0.35}
  & \textbf{97.1} / \textbf{2.95} / \textbf{0.50} \\
\bottomrule
\end{tabular}

\end{table*}

\begin{figure*}[tbh!]
  \centering
  \includegraphics[scale=0.65]{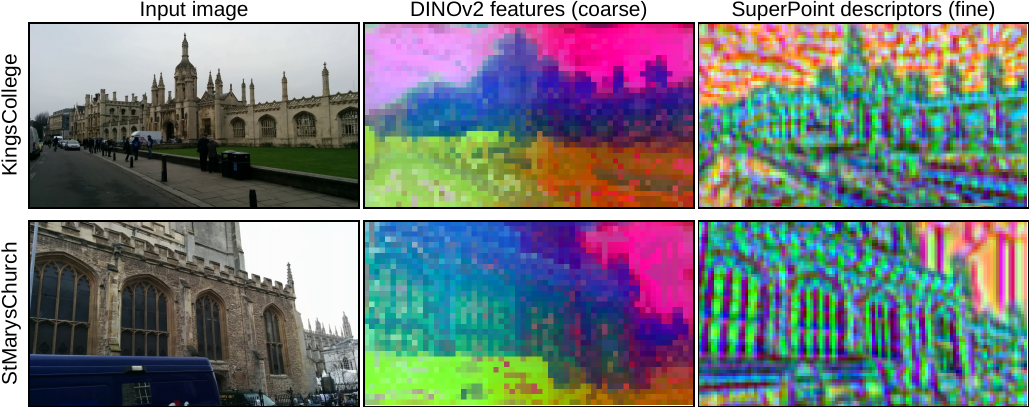}
\caption{\textbf{Coarse versus fine feature representations.} Top-3 PCA of DINOv2 backbone features (center) alongside dense SuperPoint descriptors (right), computed jointly over both query frames {(KingsCollege and StMarysChurch from Cambridge Landmarks)}.}
  \label{fig:pca}
\end{figure*}

\begin{table*}[t]
\caption{\textbf{Quantitative results on 7-Scenes.} We report
relocalization recall (\%) under the (1\,cm, 1$^\circ$) threshold,
highlighting the best entry per column in bold. {Baseline performance is}
quoted as reported in~\cite{xu2025seqace}{, except DINO-ACE (our reproduction, Sec.~S1). Methods are grouped by approximate mapping budget ({Table~S10})}.}
\label{tab:main_7scenes}
\centering
\scriptsize
\begin{tabular}{llcccccccc}
\toprule
Method & Size & Chess & Fire & Heads & Office & Pumpkin & Kitchen & Stairs & Avg. \\
\midrule
\multicolumn{10}{l}{{\textit{Non-SCR}}} \\
HLoc (SP+SG)~\cite{sarlin2019coarse,sarlin2020superglue} & $\sim$3.5\,GB
  & 96.1 & \textbf{71.2} & 82.5 & 68.6 & 75.2 & 79 & 16.3 & 69.8 \\
\midrule
\multicolumn{10}{l}{{\textit{SCR, $\sim$25-minute budget}}} \\
GLACE~\cite{wang2024glace} & 9\,MB
  & 87.4 & 67.1 & 86.3 & 60.5 & 58.7 & 75.8 & 5.8 & 63.1 \\
\midrule
\multicolumn{10}{l}{{\textit{SCR, {$\sim$5-minute} budget}}} \\
ACE~\cite{brachmann2023accelerated} & 4.1\,MB
  & 87.2 & 60.1 & 87.9 & 46.2 & 34.3 & 66.7 & 3.8 & 55.2 \\
{DINO-ACE~\cite{bruns2025ace} (ours)} & {4\,MB}
  & {53.3} & {32.0} & {61.4} & {28.5} & {17.2} & {40.5} & {0.1} & {33.3} \\ 
HSeq-Single~\cite{xu2025seqace} & 4.1\,MB
  & 94.2 & 63.5 & 87.9 & 56.2 & 51.9 & 67.7 & 11 & 61.6 \\
HSeq-Sequence~\cite{xu2025seqace} & 4.1\,MB
  & 96.5 & 68.8 & 91.7 & 59.8 & 57.8 & 71.7 & \textbf{17.4} & 66.2 \\
ReACE (ours) & 6.3\,MB
  & \textbf{98.5} & 69.3 & \textbf{92.6} & \textbf{78.2}
  & \textbf{78.1} & \textbf{83.2} & 10.9 & \textbf{73.0} \\
\bottomrule
\end{tabular}

\end{table*}

\begin{table*}[t]
\centering
\scriptsize
\caption{\textbf{Component ablation on Cambridge Landmarks.} We report the
median translation (cm) and rotation ($^\circ$) errors, highlighting the
best entry per column in bold. {The first row is quoted from~\cite{bruns2025ace}.}}
\label{tab:abl_components}
\begin{tabular}{lccccccccc}
\toprule
Configuration & \shortstack{KP-guided\\sampling} & \shortstack{Descriptor\\fusion} & \shortstack{{Tight}\\{threshold}}
  & GreatCourt & KingsCollege & OldHospital & ShopFacade & StMarysChurch & Avg. \\
\midrule
{DINO-ACE (published)} & & & & {61.4 / 0.3} & {38.9 / 0.4} & {39.8 / 0.6} & {5.3 / 0.3} & {74 / 1.8} & {43.9 / 0.68} \\
DINO-ACE {(ours)} & & & & 61.7 / 0.25 & 44.9 / 0.45 & 52.9 / 0.89 & 7.7 / 0.34 & 67.1 / 1.57 & 46.9 / 0.70 \\
 & \checkmark & & & 29.3 / 0.15 & 23.5 / 0.34 & 39.4 / 0.66 & 5.8 / 0.26 & 40.1 / 1.06 & 27.6 / 0.49 \\
 & \checkmark & \checkmark & & 26.0 / 0.12 & 18.3 / 0.25 & 27.9 / 0.50 & 5.1 / 0.19 & \textbf{30.5} / \textbf{0.89} & 21.6 / 0.39 \\
ReACE (ours) & \checkmark & \checkmark & \checkmark
  & \textbf{24.4} / \textbf{0.11} & \textbf{14.9} / \textbf{0.20} & \textbf{19.6} / \textbf{0.38}
  & \textbf{4.1} / \textbf{0.18} & 34.0 / 1.02 & \textbf{19.4} / \textbf{0.38} \\
\midrule
{SuperPoint only} & {\checkmark} & {--} & {\checkmark}
  & {61.1 / 0.19} & {36.3 / 0.50} & {1218.6 / 19.02} & {6.2 / 0.33} & {605.1 / 12.90} & {385.5 / 6.59} \\
\bottomrule
\end{tabular}
\end{table*}

\begin{table}[t]
\centering
\scriptsize
\caption{\textbf{Choice of the fine signal.} We report the median
translation (cm) and rotation ($^\circ$) errors as dataset averages,
highlighting the best entry per column in bold.}
\label{tab:abl_feature}

\begin{tabular}{llcc}
\toprule
DINOv2$+$ & Feature & 7-Scenes Avg. & Cambridge Avg. \\
\midrule
\multicolumn{2}{l}{{DINOv2 only (baseline)}} & 5.14 / 0.88 & 25.41 / 0.47 \\
\midrule
Random & $D{=}256$ & 5.65 / 0.86 & 26.01 / 0.49 \\
 & $D{=}512$ & 5.08 / 0.82 & 31.34 / 0.57 \\
{Coarse} & DINOv2 {(duplicate)} & 5.32 / 0.90 & 28.59 / 0.58 \\
\midrule
Local & DeDoDe & 1.31 / 0.39 & \textbf{18.43} / \textbf{0.35} \\
 & ELoFTR & 1.66 / 0.46 & 20.25 / 0.38 \\
 & RoMa & 1.45 / 0.42 & 20.64 / 0.40 \\
 & SuperPoint & \textbf{1.25} / \textbf{0.38} & 19.39 / 0.38 \\
 & XFeat & 2.22 / 0.54 & 19.90 / 0.36 \\
 & SIFT & 1.83 / 0.53 & 20.80 / 0.41 \\
 & RIPE & 1.33 / 0.40 & 19.87 / 0.38 \\
\bottomrule
\end{tabular}
\end{table}

\begin{table*}[t]
\centering
\scriptsize
\caption{\textbf{Single vs.\ multi-feature.} We report the median
translation (cm) and rotation ($^\circ$) errors as dataset averages,
highlighting the best entry per column in bold. Rows 2 and 5 are
ReACE and ReACE-MF (Sec.~\ref{sec_method_reace}); the DINO-ACE row
quotes its published results~\cite{bruns2025ace}.}
\label{tab:abl_single_multi}
\begin{tabular}{lcccccccc}
\toprule
 & SuperPoint & DeDoDe & ELoFTR & 7-Scenes & Cambridge & Indoor-6 & RIO10 & Map size \\
\midrule
DINO-ACE~\cite{bruns2025ace} & & & & 7.33 / -- & 43.88 / 0.68 & 5.60 / 1.02 & 83.8 / \textbf{15.9} & 4\,MB \\
\midrule
ReACE$+$ & \checkmark & & & 1.25 / 0.38 & 19.39 / 0.38
  & \textbf{2.95} / 0.50 & \textbf{55.4} / 16.1 & 6.28\,MB \\
 & \checkmark & \checkmark & & 1.14 / {0.36} & 21.50 / 0.39 & 2.99 / 0.52 & 56.7 / 16.3 & 6.91\,MB \\
 & & \checkmark & \checkmark & \textbf{1.04} / \textbf{0.33} & 19.29 / \textbf{0.37} & 3.15 / 0.53 & 57.9 / 17.1 & 6.78\,MB \\
 & \checkmark & \checkmark & \checkmark & 1.05 / \textbf{0.33} & \textbf{18.87} / \textbf{0.37}
  & 2.99 / \textbf{0.49} & {56.4 / 16.5} & 7.41\,MB \\
\bottomrule
\end{tabular}
\end{table*}

\begin{table*}[t]
\centering
\scriptsize
\caption{\textbf{Fusion-architecture ablation.} We report the median
translation (cm) and rotation ($^\circ$) errors as dataset averages and
the training time (training stage only), highlighting the best
entry per column in bold. The attention row is read against the
matched-batch control row.}
\label{tab:abl_fusion}
\begin{tabular}{llccc}
\toprule
Fusion & Method & 7-Scenes Avg. & Cambridge Avg. & Train {(min)} \\
\midrule
Concat & MLP head (ReACE default) & 1.25 / \textbf{0.38} & 19.39 / 0.38 & {4.9} \\
 & MLP head, 80$\times$128 batches (control) & \textbf{1.24} / \textbf{0.38} & 18.88 / \textbf{0.36} & {5.5} \\
\midrule
Attention & full transformer block & 1.83 / 0.52 & 32.81 / 0.59 & {6.8} \\
\midrule
Residual & frozen-base residual & 1.28 / 0.40 & \textbf{18.32} / 0.37 & {9.4} \\
 & staged residual & 4.90 / 0.86 & 21.80 / 0.43 & {9.0} \\
 & joint residual & 5.20 / 0.85 & 25.13 / 0.46 & {6.7} \\
\bottomrule
\end{tabular}
\end{table*}

\begin{table}[t]
\centering
\scriptsize
\caption{\textbf{Feature dropout.} We report the median translation (cm)
and rotation ($^\circ$) errors as dataset averages; dropout-on is the
ReACE default.}
\label{tab:abl_dropout}
\setlength{\tabcolsep}{4pt} 
\begin{tabular}{lcccc}
\toprule
 & \multicolumn{2}{c}{7-Scenes} & \multicolumn{2}{c}{Cambridge} \\
Method & drop.\ on & drop.\ off & drop.\ on & drop.\ off \\
\midrule
{DINOv2 only (baseline)} & 5.14 / 0.88 & 6.62 / 1.19 & 25.41 / 0.47 & 30.49 / 0.63 \\
ReACE & 1.25 / 0.38 & 3.58 / 0.75 & 19.39 / 0.38 & 24.21 / 0.48 \\
$+$DeDoDe & {1.21 / 0.38} & 2.23 / 0.59 & {19.00 / 0.38} & 21.98 / 0.44 \\
ReACE-MF & 1.05 / 0.33 & 1.86 / 0.52 & 18.87 / 0.37 & 18.24 / 0.38 \\
\bottomrule
\end{tabular}
\end{table}



\section{Experiments}
\label{sec:experiments}


\subsection{Experimental Setup}
\label{sec:exp_setup}

\paragraph{Baselines.}
We compare against SCR methods categorized by mapping budget. We note that mapping times are difficult to compare directly across methods, as published figures were measured on different hardware in single- or multi-GPU configurations and depend on each method's design, hyperparameters, and training schedule~\cite{brachmann2023accelerated}. We therefore group methods into a $\sim$5-minute and a $\sim$25-minute budget and match our variants to DINO-ACE: ReACE keeps DINO-ACE's training schedule and buffer size, and ReACE-Poker trains four such maps. On a single  NVIDIA RTX PRO 6000 Blackwell GPU, ReACE maps each of the twelve Cambridge and 7-Scenes scenes in 5.6--6.4\,min, against 5.4--6.1\,min for DINO-ACE (Table~S10). At query time, ReACE takes 32 and 48\,ms per image on 7-Scenes and Cambridge Landmarks, against 33 and 38\,ms for DINO-ACE (Table~S11). 

Within the $\sim$5-minute budget, we evaluate ACE~\cite{brachmann2023accelerated}, DINO-ACE~\cite{bruns2025ace}, HSeq-Single~\cite{xu2025seqace}, and ACE-G~\cite{bruns2025ace}. ReACE results are reported as average over four seeds. Within the {$\sim$25-minute} budget, we include GLACE~\cite{wang2024glace}, the 25-minute variant of ACE-G, and the ensemble methods ACE-Poker~\cite{brachmann2023accelerated} and HSeq-Ensemble~\cite{xu2025seqace}. For ReACE-Poker we follow ACE's Poker partitioning and training recipe. Beyond coordinate regressors, we evaluate the hierarchical feature-matching pipeline HLoc (SP+SG)~\cite{sarlin2019coarse,sarlin2020superglue} on Cambridge Landmarks and 7-Scenes, as well as Reloc3r~\cite{dong2025reloc3r} and MASt3R~\cite{leroy2024mast3r}+Kapture on Indoor-6 following the evaluation protocol of ACE-G. On Cambridge Landmarks and Indoor-6 we also include D2S~\cite{bui2024d2s}, which regresses scene coordinates from sparse descriptors but, unlike the regressors above that learn from posed images alone, it is supervised by a 3D SfM reconstruction of the scene and trains for hours per scene. We additionally include HSeq-Sequence where published results are available. DINO-ACE serves as our primary baseline because it shares the identical frozen DINOv2 backbone, differing from ReACE only by its dense grid sampling, absence of local descriptors, and standard pose solver. Our {implementation} reproduces the published per-scene results of DINO-ACE on 7-Scenes. {All other baseline entries are quoted from published results, with the source given in each table caption. Measured mapping times {(Table~S10)}, the Poker recipe {(Sec.~S4)}, a comparison of the two pose solvers {(Sec.~S2)}, and the DINO-ACE reproduction {(Sec.~S1)} are detailed in the Suppl. Material.}

\paragraph{Datasets and Evaluation Metrics.}
We evaluate ReACE across both indoor and outdoor environments. Primary comparisons are conducted on three standard benchmarks, including the outdoor Cambridge Landmarks dataset~\cite{kendall2015posenet} and the indoor 7-Scenes~\cite{shotton2013scene} and Indoor-6~\cite{do2022scenelandmarks} datasets. Ablation studies are additionally reported on RIO10~\cite{wald2020rio10}, whose rescan-based evaluation probes robustness to temporal scene change that ReACE does not explicitly address. Its official test split is hidden, so published comparisons are limited to validation medians without recall and are reported apart from the recall-anchored main benchmarks. ReACE nevertheless {lowers DINO-ACE's median translation error} on all four datasets (\Cref{tab:abl_single_multi}).
Following standard benchmark conventions, we report median translation (cm) and rotation (deg) errors on Cambridge Landmarks, pose recall under (25\,cm, 5$^\circ$) alongside median errors on Indoor-6, and recall under the fine (1\,cm, 1$^\circ$) threshold on 7-Scenes{, using the SfM ground-truth poses as in~\cite{brachmann2023accelerated,wang2024glace}} ({Sec.~S5}).

\paragraph{Implementation details.}
The vision backbone is the DINOv2 ViT-L/14 model with registers, following the DINO-ACE configuration of Bruns et al.~\cite{bruns2025ace}. Keypoints and local descriptors are extracted with SuperPoint~\cite{detone2018superpoint}, retaining up to 1{,}536 locations per image. Because SuperPoint descriptors have dimension $D = 256$, the scaling parameter is set to $s = \sqrt{256} = 16$ in \cref{eq_fusion}. In the default ReACE configuration, the 1024-D {DINOv2} feature sampled at keypoint locations is concatenated with the scaled 256-D SuperPoint descriptor, forming a 1280-D input to the regression head. ReACE-MF expands this input to 1792-D by incorporating coarse ELoFTR~\cite{wang2024eloftr} features and DeDoDe~\cite{edstedt2024dedode} descriptors.

{All variants inherit} the ACE one-cycle schedule {for} 25k iterations, using a 4M feature buffer and feature dropout ($p = 0.1$) at the head input~\cite{bruns2025ace}. {Our map sizes are the half-precision checkpoints on disk, while baseline map sizes are quoted as published, and storage precision may differ between methods.} At query time, \texttt{solvePnPRansac} runs for up to 1{,}000 iterations at a 10\,px inlier threshold, and the pose is then refined on its inliers at a tight 4\,px threshold. Further details are given in Sec.~S4.

\subsection{Main Results}
\label{sec:exp_main}

\paragraph{Cambridge Landmarks.}
Within the {$\sim$5-minute} budget (~\Cref{tab:main_cambridge}), ReACE reduces the average median error of its DINO-ACE baseline from 43.9\,cm and 0.68$^\circ$ to 19.4\,cm and 0.38$^\circ$, the best in this category. This advantage transfers to the {$\sim$25-minute} budget, where ReACE-Poker reaches an average error of 13.2\,cm and 0.24$^\circ$, {on par with GLACE (13.6\,cm, 0.26$^\circ$) and ahead of} all other SCR baselines {in this group}. ReACE-Poker's larger storage footprint (25.1 vs.\ 13\,MB) is offset by its trivially parallelizable ensemble mapping. {Higher accuracy is achieved only by the two methods that depend on an SfM reconstruction: HLoc, which requires a map footprint of roughly 800\,MB, and D2S, which needs hours of training per scene.} 

\Cref{fig:pca} visualizes the top three principal components of the DINOv2 patch features (center) and the dense SuperPoint descriptors (right). The backbone representations exhibit smooth semantic consistency where repeated architectural elements share similar color responses, whereas local descriptors capture high-frequency variations across individual structural instances. {The coarse feature thus identifies the semantic region, and the descriptor tells repeated instances apart.}
\paragraph{Indoor-6.}
Within the {$\sim$5-minute} mapping budget (\Cref{tab:main_indoor6}), ReACE achieves 97.1\% recall with a median translation error of 2.95\,cm, outperforming even the 25-minute variant of ACE-G. In the {$\sim$25-minute group}, ReACE-Poker is the top-performing entry across all three metrics. ReACE-Poker surpasses even the non-SCR methods despite their {map footprints of hundreds of megabytes to several gigabytes and, for MASt3R, hours of mapping time}.

\paragraph{7-Scenes.}
Under the fine (1\,cm, 1$^\circ$) threshold (\Cref{tab:main_7scenes}), ReACE achieves 73.0\% average recall, outperforming the feature-matching pipeline HLoc (69.8\%) as well as all SCR baselines, while leading on five of the seven scenes. For the remaining two scenes, Fire is led by HLoc and {the Stairs scene by} HSeq-Sequence.

\subsection{Ablation Studies and Stability Analysis}
\label{sec:ablations}

\paragraph{Isolating individual contributions.}
\Cref{tab:abl_components} evaluates the three modifications proposed by ReACE on the Cambridge Landmarks dataset. The first row reports the published DINO-ACE results obtained with the DSAC* pose solver~\cite{brachmann2021dsacstar}. To ensure a controlled comparison, the remaining rows, including a retrained DINO-ACE baseline, are evaluated with our pose solver, which yields higher error than DSAC* on Cambridge Landmarks (Sec.~S2). Each modification reduces the average error (rows 3-5), with keypoint-guided sampling contributing the largest gain. We further analyze the effect of the number of keypoints in Table~S9. Together ReACE's proposed modifications cut the DINO-ACE translation error by more than half. Finally, as a control experiment, we remove the DINOv2 features and regress coordinates solely from SuperPoint descriptors (last row). This configuration fails to converge on large outdoor scenes, supporting our hypothesis that local descriptors act as a complementary refinement signal alongside the primary semantic guidance of DINOv2 features.


\paragraph{Descriptor selection and feature concatenation.}
\Cref{tab:abl_feature} evaluates different signals fused with the {DINOv2} features to isolate the source of improvement on the 7-Scenes and Cambridge Landmarks datasets. Replacing the local descriptor with a random noise vector of equal dimension ($D=256$) yields no improvement, and leads to degradation on the Cambridge Landmarks when expanded to $D = 512$. Similarly, duplicating the {DINOv2} backbone feature as the second stream yields no gain. In contrast, fusing any local descriptor produces large improvements across both benchmarks, with SuperPoint, DeDoDe, and RIPE delivering comparable results. {The gain thus comes from the descriptor's information, not from extra input dimensions.} 

Extending this fusion to the full multi-feature configuration in~\Cref{tab:abl_single_multi} produces mixed, dataset-dependent outcomes. ReACE-MF yields marginal gains over ReACE on 7-Scenes (reducing average error from 1.25 to 1.05\,cm) and Cambridge Landmarks (improving from 19.39 to 18.87\,cm), but provides no measurable benefit on Indoor-6 or RIO10 while increasing overall model footprint. {Per-scene results for both tables are given in Tables~S4 and~S5.

\paragraph{Fusion architecture and feature dropout.}
Comparing fusion mechanisms in \Cref{tab:abl_fusion} shows that simple concatenation {is competitive with} more complex alternatives. An attention-based fusion degrades both dataset averages. 
The best residual configuration reduces translation error on Cambridge Landmarks by  $\sim1$,cm, but yields no gains on 7-Scenes while nearly doubling training time ($1.9\times$) by optimizing a second head on top of a trained ReACE model.
Applying feature dropout at the head input (\Cref{tab:abl_dropout}) substantially improves the no-descriptor baseline and ReACE. However, this effect diminishes as additional descriptors are concatenated and slightly reverses for ReACE-MF on Cambridge Landmarks, indicating that feature dropout and additional descriptors serve as partial substitutes. {Per-scene results for both tables are given in Tables~S6 and~S7.}

\paragraph{Descriptor scaling.}
Evaluating the scaling factor $s$ in \cref{eq_fusion} across $\{1, 8, 16, 32\}$ {(Tables~S2 and~S8)} shows broad insensitivity from 1 to 16 (mean discrepancy 0.18\,cm across the twelve scenes) and degraded accuracy at 32. ReACE uses the fixed $s = \sqrt{256} = 16$ convention without tuning.

\paragraph{Seed stability.}
{Each seeded experiment runs the complete pipeline with four fixed seeds and a single-threaded solver, so evaluation is deterministic.} Run-to-run variation remains minimal {(Table~S3)}. Translation standard deviations remain below 0.5,cm across 7-Scenes, Cambridge Landmarks, and Indoor-6, so our gains over DINO-ACE} exceed seed variation by {over} an order of magnitude.

\section{Conclusion}
\label{sec:conclusion}
\paragraph{Summary.}
We introduced ReACE, which revisits the role of local descriptors in scene coordinate regression. By augmenting the ACE paradigm with keypoint-guided sampling and pointwise feature fusion, ReACE achieves substantial accuracy gains across indoor and outdoor scenes, showing that coordinate regression benefits from fusing coarse semantic context with fine local geometry at the input level. ReACE preserves minutes-scale mapping and a megabyte-scale model footprint, with ablations confirming that improvements stem from injected point-level geometric cues rather than added capacity.

\paragraph{Limitations and future work.}
ReACE inherits standard coordinate regression limitations, lacking explicit disambiguation for severe perceptual aliasing like repetitive stairs and vulnerability to substantial temporal scene changes such as those in RIO10. Because our fusion operates strictly at the input level, it can serve as a drop-in representation for conditioned pretraining. Natural future directions include developing adaptive mapping-time schemes to select the optimal descriptor combination per scene, and combining our input representation with the global feature encoding and position decoder of GLACE to scale to larger scenes.

{
    \small
    \bibliographystyle{ieeenat_fullname}
    \bibliography{main}
}
\setcounter{section}{0}
\setcounter{table}{0}
\setcounter{figure}{0}
\setcounter{equation}{0}

\renewcommand{\thesection}{S\arabic{section}}
\renewcommand{\thesubsection}{S\arabic{section}.\arabic{subsection}}
\renewcommand{\thetable}{S\arabic{table}}
\renewcommand{\thefigure}{S\arabic{figure}}
\renewcommand{\theequation}{S\arabic{equation}}

\setcounter{section}{0}
\setcounter{table}{0}
\setcounter{figure}{0}
\setcounter{equation}{0}

\maketitlesupplementary
\noindent{Table, section, and figure numbers without the prefix S refer to the main paper.}

\section{{DINO-ACE Reproduction on 7-Scenes}}
\label{sec:supp_repro}

{Following prior work~\cite{brachmann2023accelerated,wang2024glace},
all our 7-Scenes results use the SfM ground-truth
poses~\cite{brachmann2021limits}. Sec.~\ref{sec:supp_gt} gives the
details.} {We compare} our DINO-ACE {implementation scene by scene with}
the published {results} (\Cref{tab:supp_repro}), using the {original} dense
sampling and {DSAC*~\cite{brachmann2021dsacstar}} solver. {With the SfM poses, it closely matches them on all seven scenes.} {The} ACE-G {paper} reports {only} median
translation on 7-Scenes, {so we compare translation only}.

\begin{table}[h]
\centering
\scriptsize
\setlength{\tabcolsep}{2.5pt}
\caption{\textbf{DINO-ACE reproduction on 7-Scenes ((median
translation, cm).}}
\label{tab:supp_repro}
\begin{tabular}{lcccccccc}
\toprule
 & Chess & Fire & Heads & Office & Pumpkin & Kitchen & Stairs & Avg. \\
\midrule
Published~\cite{bruns2025ace} & 0.9 & 1.4 & 0.8 & 1.4 & 1.8 & 1.1 & 43.9 & 7.33 \\
Ours, {SfM} & {0.96} & {1.35} & {0.85} & {1.46} & {1.74} & {1.16} & {41.04} & {6.94} \\
\bottomrule
\end{tabular}
\end{table}

\section{Solver Comparability}
\label{sec:supp_solver}

{We estimate} poses with \texttt{solvePnPRansac} on sparse
keypoints, whereas {the} published SCR {results} use {DSAC*} on a dense
{grid of predictions}. 
{When we evaluate our DINO-ACE models (one per scene) with both solvers,
the results on 7-Scenes are nearly identical outside Stairs (median
translation 1.30 vs.\ 1.25\,cm; recall at (1\,cm, 1$^\circ$) 38.6 vs.\
38.8\%). On Stairs and on Cambridge Landmarks, our solver is less accurate,
by 11.5\,cm and 3.7\,cm in median translation. The solver difference
therefore favors the published results, not ours.} {All our other results use \texttt{solvePnPRansac}; only the
DINO-ACE reproductions (\Cref{tab:supp_repro})
use DSAC*. Note also that DSAC*} {gives different
results from run to run unless it runs single-threaded} (\cref{sec:supp_impl}).

\section{{Per-Scene Results and Seed Stability}}
\label{sec:supp_perscene}

Tables.~\ref{tab:supp_feature}--\ref{tab:supp_descscale} {give per-scene
results for} every dataset-average ablation {in} the main paper
(Tables.~5--8). {All tables in this section use the same protocol} as described in the main paper. {Each cell gives} median
translation (cm)\,/\,median rotation ($^\circ$). {The last column of each
block is the average over scenes reported in} the main paper. {This section also
gives the full} descriptor-scale study (\Cref{tab:abl_descscale}) and
{the} seed-stability summary (~\Cref{tab:abl_seeds}). {On
7-Scenes (\Cref{tab:supp_descscale}), $s{=}16$ is within 0.04\,cm of
$s{=}1$ on six scenes and 2.4\,cm better on the Stairs scene, the best of the
four scales there (\Cref{tab:abl_descscale}); on Cambridge, $s{=}1$ has
the lower average (17.91 vs.\ 18.82\,cm). ReACE uses $s{=}16$.}

\begin{table}[t]
\centering
\scriptsize
\caption{\textbf{Descriptor scale.} We report the median translation (cm)
and rotation ($^\circ$) errors on the two scale-sensitive scenes{,
Stairs and StMarysChurch}.}
\label{tab:abl_descscale}
\begin{tabular}{lcc}
\toprule
Descriptor scale & Stairs & StMarysChurch \\
\midrule
1 & 8.00 / 2.09 & \textbf{29.37} / 0.98 \\
8 & 6.94 / 1.94 & 30.21 / \textbf{0.97} \\
16 (default) & \textbf{5.59} / \textbf{1.58} & 32.67 / 0.99 \\
32 & 5.84 / 1.77 & 47.58 / 1.39 \\
\bottomrule
\end{tabular}
\end{table}

\begin{table}[t]
\centering
\scriptsize
\caption{\textbf{Seed stability.} {For each seed, we average the}
median translation (cm)\,/\,rotation ($^\circ$) {over the scenes of a
dataset. Cells give the} mean$\pm$sd {of this average} over four seeds,
{so} the standard deviation is across seeds, not {scenes}. {The last row uses the Poker recipe for ReACE and ReACE-MF (Sec.~\ref{sec:supp_impl}) using four maps per scene.}}
\label{tab:abl_seeds}
\begin{tabular}{lcc}
\toprule
Dataset & ReACE & ReACE-MF \\
\midrule
7-Scenes & 1.25 / 0.381 & 1.05 / 0.325 \\
 & {$\pm$0.03} / $\pm$0.017 & $\pm$0.01 / $\pm$0.004 \\
Cambridge & 19.39 / 0.378 & 18.87 / 0.371 \\
 & $\pm$0.44 / $\pm$0.012 & {$\pm$0.25} / $\pm$0.004 \\
Indoor-6 & 2.95 / 0.496 & 2.99 / 0.488 \\
 & $\pm$0.06 / $\pm$0.008 & $\pm$0.04 / $\pm$0.011 \\
RIO10 & 55.36 / 16.14 & {56.40 / 16.52} \\
 & $\pm$1.10 / $\pm$0.46 & {$\pm$2.13 / $\pm$0.86} \\
Cambridge, Poker {($\times$4)} & 13.20 / 0.239 & 14.29 / 0.247 \\
 & $\pm$0.37 / $\pm$0.026 & $\pm$0.46 / $\pm$0.016 \\
\bottomrule
\end{tabular}
\end{table}

\begin{table*}[t]
\centering
\scriptsize
\setlength{\tabcolsep}{3pt}
\caption{\textbf{Choice of the fine signal, per scene} (expands
Table~5). Median translation (cm)\,/\,rotation
($^\circ$); best translation and rotation per column in bold. On 7-Scenes, the {baseline is within
1\,cm of every descriptor row on all scenes} except Stairs, so the
{descriptors' gain in the average comes almost entirely from the
Stairs scene}.}
\label{tab:supp_feature}
\textit{7-Scenes}\\[2pt]
\begin{tabular}{lcccccccc}
\toprule
DINOv2$+$ & Chess & Fire & Heads & Office & Pumpkin & Kitchen & Stairs & Avg. \\
\midrule
{DINOv2 only (baseline)} & 0.48 / 0.15 & 0.91 / 0.38 & 0.49 / 0.33 & 0.87 / 0.25 & 0.98 / 0.26 & 0.71 / 0.20 & 31.52 / 4.59 & 5.14 / 0.88 \\
Random $D{=}256$ & 0.47 / 0.15 & 0.88 / 0.36 & 0.46 / 0.32 & 0.97 / 0.28 & 1.08 / 0.27 & 0.71 / 0.21 & 34.99 / 4.42 & 5.65 / 0.86 \\
Random $D{=}512$ & 0.48 / 0.15 & 0.91 / 0.39 & 0.47 / 0.33 & 0.97 / 0.27 & 1.10 / 0.28 & 0.74 / 0.22 & 30.91 / 4.13 & 5.08 / 0.82 \\
DINOv2 {(duplicate)} & 0.45 / 0.15 & 0.92 / 0.38 & 0.46 / 0.32 & 0.87 / 0.26 & 0.99 / 0.25 & 0.70 / 0.20 & 32.83 / 4.74 & 5.32 / 0.90 \\
DeDoDe & 0.31 / 0.10 & 0.69 / \textbf{0.29} & 0.34 / 0.22 & 0.68 / 0.19 & 0.63 / 0.15 & 0.54 / 0.15 & 5.98 / 1.64 & 1.31 / 0.39 \\
ELoFTR & 0.33 / 0.11 & 0.78 / 0.33 & 0.35 / 0.24 & 0.74 / 0.21 & 0.64 / 0.16 & 0.55 / 0.16 & 8.23 / 2.02 & 1.66 / 0.46 \\
RoMa & \textbf{0.27} / \textbf{0.09} & \textbf{0.67} / \textbf{0.29} & \textbf{0.29} / \textbf{0.20} & 0.63 / \textbf{0.18} & 0.56 / \textbf{0.13} & \textbf{0.49} / \textbf{0.14} & 7.25 / 1.92 & 1.45 / 0.42 \\
SuperPoint & 0.28 / 0.10 & 0.72 / 0.30 & 0.30 / \textbf{0.20} & \textbf{0.61} / \textbf{0.18} & 0.57 / \textbf{0.13} & \textbf{0.49} / \textbf{0.14} & \textbf{5.75} / \textbf{1.62} & \textbf{1.25} / \textbf{0.38} \\
XFeat & 0.35 / 0.13 & 0.78 / 0.30 & 0.36 / 0.23 & 0.74 / 0.20 & 0.75 / 0.19 & 0.66 / 0.19 & 11.95 / 2.52 & 2.22 / 0.54 \\
SIFT & 0.36 / 0.12 & 0.85 / 0.36 & 0.40 / 0.28 & 0.72 / 0.21 & 0.67 / 0.16 & 0.60 / 0.17 & 9.21 / 2.39 & 1.83 / 0.53 \\
RIPE & 0.29 / 0.10 & 0.69 / \textbf{0.29} & 0.31 / 0.21 & 0.65 / 0.19 & \textbf{0.55} / 0.14 & 0.51 / 0.15 & 6.31 / 1.70 & 1.33 / 0.40 \\
\bottomrule
\end{tabular}\\[8pt]
\textit{Cambridge Landmarks}\\[2pt]
\begin{tabular}{lcccccc}
\toprule
DINOv2$+$ & GreatCourt & KingsCollege & OldHospital & ShopFacade & StMarysChurch & Avg. \\
\midrule
{DINOv2 only (baseline)} & 28.96 / 0.15 & 18.38 / 0.25 & 33.03 / 0.54 & 5.01 / 0.23 & 41.68 / 1.17 & 25.41 / 0.47 \\
Random $D{=}256$ & 31.47 / 0.18 & 23.26 / 0.32 & 30.60 / 0.52 & 5.16 / 0.22 & 39.57 / 1.19 & 26.01 / 0.49 \\
Random $D{=}512$ & 32.60 / 0.17 & 24.31 / 0.33 & 46.69 / 0.67 & 5.87 / 0.24 & 47.23 / 1.46 & 31.34 / 0.57 \\
DINOv2 {(duplicate)} & 34.36 / 0.18 & 20.51 / 0.32 & 31.53 / 0.56 & 5.35 / 0.25 & 51.18 / 1.59 & 28.59 / 0.58 \\
DeDoDe & 26.74 / 0.12 & \textbf{14.09} / 0.22 & 18.97 / \textbf{0.34} & 4.65 / 0.19 & \textbf{27.71} / \textbf{0.88} & \textbf{18.43} / \textbf{0.35} \\
ELoFTR & 28.12 / 0.13 & 15.37 / \textbf{0.19} & 20.80 / 0.42 & 4.66 / 0.19 & 32.30 / 0.97 & 20.25 / 0.38 \\
RoMa & 25.34 / 0.12 & 17.79 / 0.25 & \textbf{18.85} / 0.36 & \textbf{4.01} / \textbf{0.16} & 37.19 / 1.10 & 20.64 / 0.40 \\
SuperPoint & \textbf{24.38} / \textbf{0.11} & 14.87 / 0.20 & 19.61 / 0.38 & 4.09 / 0.18 & 34.00 / 1.02 & 19.39 / 0.38 \\
XFeat & 28.93 / 0.13 & 16.00 / 0.20 & 19.88 / 0.36 & 4.52 / 0.21 & 30.17 / 0.90 & 19.90 / 0.36 \\
SIFT & 27.21 / 0.12 & 15.55 / 0.21 & 21.45 / 0.40 & 4.81 / 0.21 & 35.00 / 1.09 & 20.80 / 0.41 \\
RIPE & 25.32 / 0.12 & 15.77 / 0.21 & 19.95 / 0.38 & 4.25 / 0.19 & 34.05 / 1.01 & 19.87 / 0.38 \\
\bottomrule
\end{tabular}
\end{table*}

\begin{table*}[t]
\centering
\scriptsize
\setlength{\tabcolsep}{3pt}
\caption{\textbf{Single vs.\ multi-feature, per scene} (expands
Table 6). Rows as in the main table: ReACE{; ReACE with DeDoDe added}
($+$DeDoDe){; DINOv2 with DeDoDe and ELoFTR but no SuperPoint
descriptor} (DeDoDe$+$ELoFTR){; and} ReACE-MF. The
{meter}-level errors on RIO10 Scenes 04 and 08 affect every method,
including the published baselines, which they arise from the change between
the mapping and validation scans of that benchmark.}
\label{tab:supp_single_multi}
\textit{7-Scenes}\\[2pt]
\begin{tabular}{lcccccccc}
\toprule
 & Chess & Fire & Heads & Office & Pumpkin & Kitchen & Stairs & Avg. \\
\midrule
ReACE & 0.28 / 0.10 & 0.72 / 0.30 & \textbf{0.30} / \textbf{0.20} & \textbf{0.61} / \textbf{0.18} & 0.57 / \textbf{0.13} & \textbf{0.49} / \textbf{0.14} & 5.75 / 1.62 & 1.25 / 0.38 \\
$+$DeDoDe & \textbf{0.27} / \textbf{0.09} & \textbf{0.70} / \textbf{0.29} & 0.31 / 0.21 & 0.62 / \textbf{0.18} & \textbf{0.56} / \textbf{0.13} & 0.51 / \textbf{0.14} & 4.99 / 1.50 & 1.14 / 0.36 \\
DeDoDe$+$ELoFTR & 0.30 / 0.10 & \textbf{0.70} / 0.30 & 0.33 / 0.23 & 0.68 / 0.19 & 0.65 / 0.16 & 0.54 / 0.15 & \textbf{4.08} / \textbf{1.18} & \textbf{1.04} / \textbf{0.33} \\
ReACE-MF & 0.28 / \textbf{0.09} & \textbf{0.70} / 0.30 & 0.33 / 0.21 & 0.63 / \textbf{0.18} & 0.59 / 0.14 & \textbf{0.49} / \textbf{0.14} & 4.32 / 1.21 & 1.05 / \textbf{0.33} \\
\bottomrule
\end{tabular}\\[8pt]
\textit{Cambridge Landmarks}\\[2pt]
\begin{tabular}{lcccccc}
\toprule
 & GreatCourt & KingsCollege & OldHospital & ShopFacade & StMarysChurch & Avg. \\
\midrule
ReACE & 24.38 / \textbf{0.11} & 14.87 / 0.20 & 19.61 / 0.38 & \textbf{4.09} / \textbf{0.18} & 34.00 / 1.02 & 19.39 / 0.38 \\
$+$DeDoDe & 29.54 / 0.12 & \textbf{13.80} / \textbf{0.19} & 19.92 / 0.35 & 4.49 / 0.20 & 39.76 / 1.10 & 21.50 / 0.39 \\
DeDoDe$+$ELoFTR & 24.91 / 0.13 & 16.20 / 0.22 & \textbf{18.88} / \textbf{0.34} & 4.32 / \textbf{0.18} & 32.12 / 1.00 & 19.29 / \textbf{0.37} \\
ReACE-MF & \textbf{23.16} / 0.12 & 15.23 / 0.21 & 20.08 / 0.37 & 4.34 / 0.19 & \textbf{31.52} / \textbf{0.97} & \textbf{18.87} / \textbf{0.37} \\
\bottomrule
\end{tabular}\\[8pt]
\textit{Indoor-6}\\[2pt]
\begin{tabular}{lccccccc}
\toprule
 & Scene 1 & Scene 2a & Scene 3 & Scene 4a & Scene 5 & Scene 6 & Avg. \\
\midrule
ReACE & \textbf{2.91} / \textbf{0.50} & \textbf{3.82} / \textbf{0.40} & 2.97 / 0.57 & \textbf{2.24} / \textbf{0.51} & \textbf{3.99} / 0.64 & 1.78 / \textbf{0.35} & \textbf{2.95} / 0.50 \\
$+$DeDoDe & 2.97 / \textbf{0.50} & 3.87 / 0.43 & \textbf{2.78} / 0.54 & 2.59 / 0.59 & 4.00 / 0.66 & \textbf{1.74} / 0.37 & 2.99 / 0.52 \\
DeDoDe$+$ELoFTR & 3.04 / 0.52 & 4.12 / 0.41 & 3.16 / 0.61 & 2.44 / 0.57 & 4.27 / 0.68 & 1.86 / 0.38 & 3.15 / 0.53 \\
ReACE-MF & 2.98 / \textbf{0.50} & 3.94 / \textbf{0.40} & 2.88 / \textbf{0.51} & 2.28 / 0.55 & 4.08 / \textbf{0.61} & 1.80 / 0.36 & 2.99 / \textbf{0.49} \\
\bottomrule
\end{tabular}\\[8pt]
\textit{RIO10}\\[2pt]
\begin{tabular}{lccccccccccc}
\toprule
 & Sc.\,01 & Sc.\,02 & Sc.\,03 & Sc.\,04 & Sc.\,05 & Sc.\,06 & Sc.\,07 & Sc.\,08 & Sc.\,09 & Sc.\,10 & Avg. \\
\midrule
ReACE & 28.40 / 9.04 & 17.55 / 5.38 & 17.64 / 5.01 & \textbf{151.2} / \textbf{53.2} & 28.96 / 11.3 & 9.47 / 3.49 & 17.39 / 4.87 & 179.7 / 29.2 & 51.43 / 16.9 & 51.82 / 23.0 & \textbf{55.36} / \textbf{16.1} \\
$+$DeDoDe & \textbf{25.43} / 8.72 & 15.81 / 4.91 & \textbf{16.27} / 4.87 & 162.1 / 55.0 & \textbf{26.24} / \textbf{10.5} & 8.59 / 3.31 & \textbf{16.78} / \textbf{4.78} & 188.1 / 29.9 & 57.76 / 18.8 & \textbf{50.26} / \textbf{22.5} & 56.73 / 16.3 \\
DeDoDe$+$ELoFTR & 27.77 / 9.08 & 16.66 / 5.22 & 17.67 / 5.05 & 166.2 / 60.6 & 29.76 / 10.9 & \textbf{8.22} / 2.99 & 19.21 / 5.14 & 192.3 / 32.2 & \textbf{48.98} / \textbf{16.3} & 51.85 / 23.8 & 57.86 / 17.1 \\
ReACE-MF & 25.61 / \textbf{8.45} & \textbf{15.61} / \textbf{4.86} & 16.55 / \textbf{4.65} & 170.5 / 57.2 & {31.46 / 12.3} & 8.44 / \textbf{2.96} & 18.12 / 5.00 & \textbf{172.4} / \textbf{28.9} & 53.22 / 17.6 & 52.13 / 23.2 & {56.40 / 16.5} \\
\bottomrule
\end{tabular}
\end{table*}

\begin{table*}[t]
\centering
\scriptsize
\setlength{\tabcolsep}{3pt}
\caption{\textbf{Fusion architecture, per scene} (expands
Table~7).}
\label{tab:supp_fusion}
\textit{7-Scenes}\\[2pt]
\begin{tabular}{lcccccccc}
\toprule
Fusion & Chess & Fire & Heads & Office & Pumpkin & Kitchen & Stairs & Avg. \\
\midrule
MLP head (ReACE default) & 0.28 / 0.10 & 0.72 / \textbf{0.30} & \textbf{0.30} / \textbf{0.20} & 0.61 / 0.18 & 0.57 / \textbf{0.13} & 0.49 / 0.14 & \textbf{5.75} / \textbf{1.62} & 1.25 / \textbf{0.38} \\
MLP head, 80$\times$128 {batches} (control) & 0.27 / 0.10 & \textbf{0.71} / \textbf{0.30} & \textbf{0.30} / \textbf{0.20} & 0.57 / 0.17 & 0.53 / \textbf{0.13} & 0.46 / \textbf{0.13} & 5.82 / 1.64 & \textbf{1.24} / \textbf{0.38} \\
full transformer block & \textbf{0.26} / \textbf{0.09} & 0.73 / 0.33 & 0.31 / 0.21 & \textbf{0.55} / \textbf{0.16} & \textbf{0.51} / \textbf{0.13} & \textbf{0.45} / \textbf{0.13} & 9.97 / 2.59 & 1.83 / 0.52 \\
frozen-base residual & 0.28 / 0.10 & 0.74 / \textbf{0.30} & 0.32 / 0.21 & 0.62 / 0.18 & 0.55 / \textbf{0.13} & 0.48 / \textbf{0.13} & 5.94 / 1.72 & 1.28 / {0.40} \\
staged residual & 0.43 / 0.15 & 0.90 / 0.37 & 0.46 / 0.31 & 0.85 / 0.24 & 0.97 / 0.26 & 0.68 / 0.20 & 29.99 / 4.51 & 4.90 / 0.86 \\
joint residual & 0.45 / 0.15 & 0.90 / 0.37 & 0.49 / 0.33 & 0.85 / 0.24 & 1.00 / 0.26 & 0.69 / 0.20 & 32.01 / 4.37 & 5.20 / 0.85 \\
\bottomrule
\end{tabular}\\[8pt]
\textit{Cambridge Landmarks}\\[2pt]
\begin{tabular}{lcccccc}
\toprule
Fusion & GreatCourt & KingsCollege & OldHospital & ShopFacade & StMarysChurch & Avg. \\
\midrule
MLP head (ReACE default) & 24.38 / 0.11 & 14.87 / 0.20 & 19.61 / 0.38 & 4.09 / \textbf{0.18} & 34.00 / 1.02 & 19.39 / 0.38 \\
MLP head, 80$\times$128 {batches} (control) & {25.51} / 0.10 & \textbf{12.54} / \textbf{0.19} & 20.64 / \textbf{0.36} & 4.13 / \textbf{0.18} & \textbf{31.58} / \textbf{0.95} & {18.88} / \textbf{0.36} \\
full transformer block & {\textbf{13.02}} / \textbf{0.07} & 21.67 / 0.28 & 29.43 / 0.54 & \textbf{3.84} / 0.20 & 96.09 / 1.87 & {32.81} / 0.59 \\
frozen-base residual & 22.05 / 0.11 & 13.38 / \textbf{0.19} & \textbf{19.60} / 0.38 & 4.27 / \textbf{0.18} & 32.29 / 0.97 & \textbf{18.32} / 0.37 \\
staged residual & 26.79 / 0.14 & 17.63 / 0.27 & 26.44 / 0.41 & 4.87 / 0.22 & 33.27 / 1.09 & 21.80 / 0.43 \\
joint residual & 30.01 / 0.15 & 18.29 / 0.25 & 31.25 / 0.53 & 4.67 / 0.20 & 41.44 / 1.19 & 25.13 / 0.46 \\
\bottomrule
\end{tabular}
\end{table*}

\begin{table*}[t]
\centering
\scriptsize
\setlength{\tabcolsep}{3pt}
\caption{\textbf{Feature dropout, per scene} (expands
Table~8). {The two rows of each pair differ only in dropout. Both
$+$DeDoDe rows therefore use} $s{=}16${, which keeps the descriptor
magnitude of ReACE. Their} dropout-on cells {thus} differ from the
$+$DeDoDe row of Table~6{, which uses $s{=}\sqrt{512}$}. {Dropout helps most} where errors are
large {(}Stairs and StMarysChurch{), and its benefit} shrinks toward
zero as {more} descriptors are added {to ReACE}.}
\label{tab:supp_dropout}
\textit{7-Scenes}\\[2pt]
\begin{tabular}{llcccccccc}
\toprule
Method & Dropout & Chess & Fire & Heads & Office & Pumpkin & Kitchen & Stairs & Avg. \\
\midrule
{DINOv2 only (baseline)} & on & 0.48 / 0.15 & 0.91 / 0.38 & 0.49 / 0.33 & 0.87 / 0.25 & 0.98 / 0.26 & 0.71 / 0.20 & 31.52 / 4.59 & 5.14 / 0.88 \\
{DINOv2 only (baseline)} & off & 0.51 / 0.17 & 1.26 / 0.49 & 0.54 / 0.37 & 0.90 / 0.26 & 1.20 / 0.30 & 0.80 / 0.22 & 41.15 / 6.55 & 6.62 / 1.19 \\
ReACE & on & 0.28 / 0.10 & 0.72 / 0.30 & 0.30 / 0.20 & 0.61 / 0.18 & 0.57 / 0.13 & 0.49 / 0.14 & 5.75 / 1.62 & 1.25 / 0.38 \\
ReACE & off & 0.29 / 0.10 & 0.79 / 0.32 & 0.34 / 0.23 & 0.63 / 0.18 & 0.57 / 0.14 & 0.52 / 0.14 & 21.90 / 4.17 & 3.58 / 0.75 \\
$+$DeDoDe & on & 0.28 / 0.09 & 0.70 / 0.30 & 0.32 / 0.20 & 0.63 / 0.18 & 0.58 / 0.14 & 0.49 / 0.14 & 5.46 / 1.63 & 1.21 / 0.38 \\
$+$DeDoDe & off & 0.29 / 0.10 & 0.80 / 0.33 & 0.33 / 0.22 & 0.65 / 0.19 & 0.64 / 0.16 & 0.53 / 0.15 & 12.40 / 3.01 & 2.23 / 0.59 \\
ReACE-MF & on & 0.28 / 0.09 & 0.70 / 0.30 & 0.33 / 0.21 & 0.63 / 0.18 & 0.59 / 0.14 & 0.49 / 0.14 & 4.32 / 1.21 & 1.05 / 0.33 \\
ReACE-MF & off & 0.28 / 0.10 & 0.80 / 0.33 & 0.33 / 0.22 & 0.66 / 0.19 & 0.64 / 0.15 & 0.52 / 0.15 & 9.78 / 2.51 & 1.86 / 0.52 \\
\bottomrule
\end{tabular}\\[8pt]
\textit{Cambridge Landmarks}\\[2pt]
\begin{tabular}{llcccccc}
\toprule
Method & Dropout & GreatCourt & KingsCollege & OldHospital & ShopFacade & StMarysChurch & Avg. \\
\midrule
{DINOv2 only (baseline)} & on & 28.96 / 0.15 & 18.38 / 0.25 & 33.03 / 0.54 & {5.01 / 0.23} & {41.68 / 1.17} & {25.41} / 0.47 \\
{DINOv2 only (baseline)} & off & 29.45 / 0.15 & 21.44 / 0.30 & 31.61 / 0.57 & 6.38 / 0.33 & 63.59 / 1.77 & 30.49 / 0.63 \\
ReACE & on & 24.38 / 0.11 & 14.87 / 0.20 & 19.61 / 0.38 & 4.09 / 0.18 & 34.00 / 1.02 & 19.39 / 0.38 \\
ReACE & off & 22.50 / 0.11 & 15.40 / 0.21 & 25.91 / 0.46 & 4.04 / 0.20 & 53.20 / 1.44 & 24.21 / 0.48 \\
$+$DeDoDe & on & 25.93 / 0.12 & 13.40 / 0.20 & 17.23 / 0.37 & 4.29 / 0.20 & 34.14 / 1.03 & 19.00 / 0.38 \\
$+$DeDoDe & off & 22.99 / 0.11 & 15.05 / 0.20 & 23.66 / 0.44 & 4.46 / 0.19 & 43.73 / 1.26 & 21.98 / 0.44 \\
ReACE-MF & on & 23.16 / 0.12 & 15.23 / 0.21 & 20.08 / 0.37 & 4.34 / 0.19 & 31.52 / 0.97 & 18.87 / 0.37 \\
ReACE-MF & off & 20.26 / 0.11 & 14.49 / 0.20 & 20.35 / 0.39 & 4.43 / 0.18 & 31.69 / 0.99 & 18.24 / 0.38 \\
\bottomrule
\end{tabular}
\end{table*}

\begin{table*}[t]
\centering
\scriptsize
\setlength{\tabcolsep}{3pt}
\caption{\textbf{Descriptor scale, per scene{.}} {$s{=}1$ vs.\
$s{=}16$ on all twelve scenes} (Sec.~4.3).} 
\label{tab:supp_descscale}
\textit{7-Scenes}\\[2pt]
\begin{tabular}{lcccccccc}
\toprule
Scale & Chess & Fire & Heads & Office & Pumpkin & Kitchen & Stairs & Avg. \\
\midrule
$s{=}1$ & \textbf{0.28} / \textbf{0.10} & \textbf{0.70} / 0.31 & 0.33 / 0.22 & \textbf{0.62} / \textbf{0.18} & \textbf{0.56} / 0.14 & \textbf{0.46} / 0.14 & 8.00 / 2.09 & 1.56 / 0.45 \\
$s{=}16$ (default) & \textbf{0.28} / \textbf{0.10} & 0.74 / \textbf{0.30} & \textbf{0.31} / \textbf{0.21} & \textbf{0.62} / \textbf{0.18} & 0.57 / \textbf{0.13} & 0.48 / \textbf{0.13} & \textbf{5.59} / \textbf{1.58} & \textbf{1.23} / \textbf{0.38} \\
\bottomrule
\end{tabular}\\[8pt]
\textit{Cambridge Landmarks}\\[2pt]
\begin{tabular}{lcccccc}
\toprule
Scale & GreatCourt & KingsCollege & OldHospital & ShopFacade & StMarysChurch & Avg. \\
\midrule
$s{=}1$ & \textbf{23.43} / \textbf{0.11} & 14.81 / 0.20 & \textbf{17.91} / \textbf{0.33} & \textbf{4.03} / \textbf{0.18} & \textbf{29.37} / \textbf{0.98} & \textbf{17.91} / \textbf{0.36} \\
$s{=}16$ (default) & 23.83 / \textbf{0.11} & \textbf{14.02} / \textbf{0.19} & 19.48 / 0.37 & 4.13 / \textbf{0.18} & 32.67 / 0.99 & 18.82 / 0.37 \\
\bottomrule
\end{tabular}
\end{table*}

\section{Detailed Implementation and Model Specifications}
\label{sec:supp_impl}

\paragraph{Architecture.}
{We use ACE's MLP head}~\cite{brachmann2023accelerated} {with}
512-unit hidden layers {and} one residual block. {We change} only its
input layer, which {we widen} to the fused dimension (1280-D for ReACE,
1792-D for ReACE-MF). {The head outputs} homogeneous coordinates{,}
with the {homogeneous component} clamped to $[0.01, 4.0]$. {We resize
images} to a short side of 480\,px{. The} backbone is DINOv2 ViT-L/14 with
registers~\cite{oquab2023dinov2}, and {we sample} its patch grid
{bilinearly} at the keypoint locations.

\paragraph{Feature extraction.}
SuperPoint~\cite{detone2018superpoint} {keeps} at most 1{,}536
keypoints per image, {with a non-maximum suppression} radius of 4\,px,
{a} detection threshold {of} 0.005, a 4\,px border margin, and no
sub-pixel refinement. ReACE-MF adds {two features:} the DeDoDe-B
descriptor~\cite{edstedt2024dedode} {at} the SuperPoint {keypoints},
and ELoFTR's~\cite{wang2024eloftr} coarse-level feature map (256-D at $1/8$
resolution), {which we sample bilinearly like the DINOv2 grid. We
multiply the ELoFTR features by} 1.34 {so that their per-element
magnitude matches that of the DINOv2 features. 1.34 is} the {ratio of the
two} per-element RMS {values (0.72 and 0.54),}
measured once on {30 images from GreatCourt and Stairs}. All extractors
use {the} published weights {and are kept frozen}.

\paragraph{{Number of keypoints.}}
{In \Cref{tab:abl_kpthr}, we evaluate the SuperPoint detection
threshold multiplied and divided by ten on the Cambridge Landmark and 7Scenes datasets. A higher threshold keeps fewer
keypoints per image, and a lower one keeps more. With the lower threshold,
26--100\% of the mapping images reach the cap of 1{,}536 keypoints, depending
on the scene. Neither setting lowers the average error. On Stairs,
both settings raise the error, by 1.7\,cm ($\times$10) and 0.9\,cm
($\div$10), which is four to eight times the seed standard deviation of the
default. On the other six 7-Scenes scenes, the error changes by at most
0.05\,cm. On Cambridge, the average rises by 0.5\,cm ($\times$10) and
0.9\,cm ($\div$10), about one and two seed standard deviations.}

\begin{table}[t]
\centering
\scriptsize
\caption{{\textbf{Number of keypoints.} SuperPoint detection threshold
multiplied and divided by ten. Keypoints: mean per mapping image. Errors:
median translation (cm) / rotation ($^\circ$), averaged over the scenes of
each dataset.}}
\label{tab:abl_kpthr}
\setlength{\tabcolsep}{2.8pt}
\begin{tabular}{lccccc}
\toprule
 & \multicolumn{2}{c}{{Keypoints}} & \multicolumn{3}{c}{{Median error}} \\
\cmidrule(lr){2-3}\cmidrule(lr){4-6}
{Threshold} & {7-Sc.} & {Camb.} & {7-Scenes} & {Stairs} & {Cambridge} \\
\midrule
{0.05 ($\times$10)} & {237} & {727} & {1.47 / 0.47} & {7.43 / 2.30} & {19.86 / 0.40} \\
{0.005 (default)} & {404} & {1{,}122} & {\textbf{1.25} / \textbf{0.38}} & {\textbf{5.75} / \textbf{1.62}} & {\textbf{19.39} / \textbf{0.38}} \\
{0.0005 ($\div$10)} & {792} & {1{,}425} & {1.40 / 0.42} & {6.65 / 1.84} & {20.29 / \textbf{0.38}} \\
\bottomrule
\end{tabular}
\end{table}

\paragraph{ReACE-MF configuration.}
ReACE-MF {uses the same} detector, {head} architecture, solver parameters, and 25k-iteration training budget {as} ReACE{. The difference is the per-point features used}. The 1792-D head input {concatenates four features:} the 1024-D DINOv2 feature, the 256-D coarse-level ELoFTR feature~\cite{wang2024eloftr}, the 256-D SuperPoint descriptor~\cite{detone2018superpoint}, and the 256-D DeDoDe descriptor~\cite{edstedt2024dedode}. {We sample the DINOv2 and ELoFTR feature maps separately at each keypoint, each at its own resolution, and concatenate the two samples.} {We $\ell_2$-normalize the SuperPoint and DeDoDe descriptors} jointly {over} their combined 512 dimensions. {We keep} $s = 16$ {instead of} $\sqrt{512}$, {so the descriptors enter the head at the same total magnitude as in ReACE and only their content changes}.

\paragraph{Training.} All models {follow ACE's optimization}~\cite{brachmann2023accelerated}{: a} clamped reprojection loss {whose threshold shrinks during training}, homogeneous {coordinates}, and {the} MLP head. {We train for 25k iterations with} AdamW {at batch size 5{,}120 in half
precision, using} ACE's one-cycle schedule{. The learning rate starts at
$1/25$ of its peak of} $5{\times}10^{-3}$ and {decays} to {$10^{-4}$ of the
starting rate}. The buffer {holds} 4M keypoint samples{,} collected once from the mapping images {and shuffled}{. We apply} feature dropout ($p{=}0.1$, no
additive noise) {to} the fused head input {(ablated in Sec.~4.3)}.

\paragraph{{Mapping time.}}
{Mapping times are difficult to compare exactly. They depend on
hardware, implementation, and hyper-parameters, and the published
numbers were measured by their authors on different GPUs. The ACE authors
make the same remark~\cite{brachmann2023accelerated}. The main tables
therefore only group methods by approximate mapping budget.
~\Cref{tab:supp_runtime} lists the published times and the hardware
they were measured on.

To check the {$\sim$5-minute} budget on our hardware, we ran ACE
and DINO-ACE with their default settings on ShopFacade, on an otherwise idle machine. ACE takes 3.9\,min (published:
5\,min) and DINO-ACE takes 5.3\,min (published: 5\,min). With the same
settings, iteration count, and buffer size, ReACE takes 5.8\,min,
{about 30\,s} (9\%) more than DINO-ACE: 11\,s more for keypoint and
descriptor extraction and 19\,s more for training the wider head input.
{Over all twelve Cambridge and 7-Scenes scenes (one run each), ReACE takes 5.6--6.4\,min and DINO-ACE 5.4--6.1\,min. ReACE is 1--9\% slower on every scene. Times vary little across scenes because the iteration count and buffer size are fixed.}
ReACE-MF takes 8.9\,min, mostly because of its additional feature
extractors. {All our times cover feature and descriptor extraction, buffer construction, and training, on one NVIDIA RTX PRO 6000 with cuDNN enabled.}}

\begin{table*}[t]
\centering
\scriptsize
\setlength{\tabcolsep}{6pt}
\caption{\textbf{{Mapping times.}} {Published mapping times with
the hardware reported by their authors, and our full mapping times on
one NVIDIA RTX PRO 6000. ``--'': not stated or not measured.
$^\dagger$COLMAP triangulation times estimated
in~\cite{brachmann2023accelerated} (Cambridge / 7-Scenes); for Cambridge
we use its 12-Scenes estimate. $^\ddagger$As listed in~\cite{bruns2025ace}.
$^\ast$7-Scenes on one GPU / Cambridge on four GPUs. $^\S$ShopFacade on an
otherwise idle machine.
$^\P$ACE-G states a single GPU only for its default 5-minute variant.
$^\|$Range over the five Cambridge and seven 7-Scenes scenes, one run per scene on an idle GPU while the machine ran other jobs.
$^\#$Four times the single-map time (computed, not measured).}}
\label{tab:supp_runtime}
\begin{tabular}{lllll}
\toprule
Method & Published & {Mapping hardware} & {Ours} & {Ours, 12 scenes$^\|$} \\
\midrule
HLoc (SP+SG)~\cite{sarlin2019coarse} & {$\approx$35\,min / 3.3\,h$^\dagger$} & -- & -- & {--} \\
{D2S~\cite{bui2024d2s}} & {9.4\,h} & {GTX 1080 Ti} & -- & {--} \\
{MASt3R~\cite{leroy2024mast3r}+Kapture} & {5--10\,h$^\ddagger$} & -- & -- & {--} \\
{Reloc3r~\cite{dong2025reloc3r}} & {$<$5\,min$^\ddagger$} & -- & -- & {--} \\
\midrule
GLACE~\cite{wang2024glace} & {6 / 20\,min$^\ast$} & {Quadro RTX 6000} & -- & {--} \\
{ACE-Poker ($\times$4)~\cite{brachmann2023accelerated}} & {20\,min} & {V100} & -- & {--} \\
{HSeq-Ensemble ($\times$4)~\cite{xu2025seqace}} & {4$\times$7\,min} & {RTX 3090} & -- & {--} \\
ACE-G (25-min)~\cite{bruns2025ace} & 25\,min & {not stated$^\P$; pre-training: 8$\times$A100} & {--} & {--} \\
\midrule
ACE~\cite{brachmann2023accelerated} & 5\,min & {V100} & 3.9\,min{$^\S$} & {--} \\
DINO-ACE~\cite{bruns2025ace} & 5\,min & {--} & {5.3\,min$^\S$} & {5.4--6.1\,min} \\
HSeq-Single~\cite{xu2025seqace} & 7\,min & {RTX 3090} & -- & {--} \\
ACE-G (5-min)~\cite{bruns2025ace} & {$\approx$5\,min} & {single GPU, model not stated; pre-training: 8$\times$A100} & {--} & {--} \\
\midrule
ReACE & -- & -- & {5.8\,min$^\S$} & {5.6--6.4\,min} \\
ReACE-MF & -- & -- & {8.9\,min$^\S$} & {7.4--9.0\,min} \\
ReACE-Poker ($\times$4) & -- & -- & {4$\times$5.8\,min$^\#$} & {4$\times$(5.6--6.4)\,min$^\#$} \\
\bottomrule
\end{tabular}
\end{table*}

\paragraph{{Query time.}}
{We time relocalization on the same GPU with cuDNN enabled, from the
loaded image to the estimated pose, skipping the first 20 images as
warm-up and keeping the faster of two runs. Averaged over scenes, ReACE
takes 32\,ms per image on 7-Scenes and 48\,ms on Cambridge Landmarks,
against 33 and 38\,ms for DINO-ACE with DSAC*. ReACE-MF takes 47 and
59\,ms. ReACE-Poker relocalizes each image with all four maps, taking 332\,ms on Cambridge Landmarks; {this is an upper bound, because our timing recomputes the image features for each map}. \Cref{tab:supp_query} gives per-scene times.}

\begin{table*}[t]
\centering
\scriptsize

\setlength{\tabcolsep}{3pt}
\caption{\textbf{Query time per scene} (ms per image).}
\label{tab:supp_query}
\begin{tabular}{lcccccccc}
\toprule
Method & Chess & Fire & Heads & Office & Pumpkin & Kitchen & Stairs & Avg. \\
\midrule
DINO-ACE & 25.1 & 40.6 & 30.3 & 26.9 & 29.9 & 34.5 & 40.9 & 32.6 \\
ReACE & 22.6 & 29.2 & 25.6 & 22.5 & 23.1 & 27.5 & 70.0 & 31.5 \\
ReACE-MF & 38.5 & 42.9 & 40.8 & 38.1 & 38.8 & 43.1 & 83.4 & 46.5 \\
\bottomrule
\end{tabular}\\[8pt]
\begin{tabular}{lcccccc}
\toprule
Method & GreatCourt & KingsCollege & OldHospital & ShopFacade & StMarysChurch & Avg. \\
\midrule
DINO-ACE & 48.5 & 29.0 & 37.7 & 27.4 & 47.6 & 38.0 \\
ReACE & 33.4 & 37.4 & 51.9 & 30.8 & 88.3 & 48.4 \\
ReACE-MF & 49.0 & 51.6 & 52.9 & 44.5 & 98.3 & 59.3 \\
ReACE-Poker ($\times$4) & 272.9 & 335.1 & 328.8 & 346.2 & 376.3 & 331.9 \\
\bottomrule
\end{tabular}
\end{table*}

\paragraph{Pose solver and determinism.}
{Pose hypotheses come from} OpenCV's \texttt{solvePnPRansac} with {up to} 1{,}000
iterations {and} a 10\,px {inlier} threshold. {The pose is then iteratively refined on its inliers at 4\,px until convergence {(10\,px in the rows of Tab.~4 without the tight threshold)}.} {The 4\,px
value follows the HSeq implementation~\cite{xu2025seqace}.} {RANSAC is
random, and multi-threaded solvers can return different poses from run to
run. All evaluations therefore fix the solver seed and run the solver
single-threaded.}

\paragraph{Poker ensembles.}
{The ReACE-Poker ($\times$4) rows} follow ACE~\cite{brachmann2023accelerated}{.
We split the} mapping frames {into} four spatial clusters {with} bisecting
$k$-means on {the} camera centers {and train one map per cluster. Each
map localizes every query independently, and we keep the pose with the most
RANSAC inliers. Map sizes are the sum over the four heads.}

\paragraph{Map sizes.}
Our sizes are {those of the half-precision} checkpoints on disk{, in MB}.
{We quote published sizes as printed, in each paper's own precision. ACE
and our models store the MLP head in half precision, whereas ACE-G's 12\,MB are its map codes
($4096{\times}768$ values) at full precision. We note that the majority of related papers do not
state the precision, so map sizes are therefore only roughly comparable.}

\paragraph{{Seed protocol.}}
{ReACE, ReACE-MF, and ReACE-Poker results are means over four pipeline
seeds wherever they appear. Each seed repeats training, localization, and
evaluation. All other cells are single runs with the default seed. Where
we also ran four seeds for an ablation cell, we report their mean.
\Cref{tab:abl_seeds} shows that seed variation is small compared
with the gains we report.} The training times of Table~7 are wall-clock of
the training stage only (buffer and setup excluded), {measured on ShopFacade with cuDNN enabled}.

\paragraph{Per-keypoint error figure.}
{For each keypoint, the error map in} Fig.~1 {shows the} distance
{from} the predicted scene coordinate to the ground-truth viewing ray{.
Because the test frames have no depth, this distance is} a lower bound on the
3D error. Each colored cell is one $14{\times}14$\,px patch of the DINOv2
grid{, colored by the error of its sampled keypoint. Many keypoints lie on
content that cannot be mapped, such as transient objects and vegetation, so
the median error over all keypoints is high. The pose depends only on the
accurate keypoints that RANSAC selects. The three models in Fig.~1
(SuperPoint only, DINOv2 only, and ReACE) use the same keypoints and
evaluation protocol}.

\section{{SfM and D-SLAM Poses on 7-Scenes}}
\label{sec:supp_gt}

{7-Scenes comes with two sets of ground-truth poses: the original
depth-SLAM (D-SLAM) trajectories~\cite{shotton2013scene} and poses from
SfM~\cite{brachmann2021limits}. Prior work reports results on both sets,
because a relocalizer can score better against one set than against the
other~\cite{brachmann2023accelerated,wang2024glace}. We use the SfM poses because the published DINO-ACE results match our
evaluation against them, not against the D-SLAM poses. With SfM poses our implementation closesly matches the published results on all seven scenes.
With D-SLAM poses, its error is 1.4--2.9$\times$ as large on six of the
seven scenes.}

\end{document}